\documentclass{article}

\PassOptionsToPackage{numbers,sort&compress}{natbib}
\usepackage[preprint]{neurips_2019}
\usepackage{amsmath}
\usepackage{algorithm}
\usepackage{algorithmic}
\usepackage[utf8]{inputenc} 
\usepackage[T1]{fontenc}    
\usepackage{hyperref}       
\usepackage{url}            
\usepackage{booktabs}       
\usepackage{amsfonts}       
\usepackage{nicefrac}       
\usepackage{microtype}      
\usepackage{graphicx}
\graphicspath{{graphics/}}
\newcommand{\methodKNN}{Kernelised Neural Network} 
\title{Training Neural Networks using SAT solvers}

\author{%
  Subham Sekhar Sahoo \\
  Department of Electrical Engineering\\
  Indian Institute of Technology - Kharagpur\\
  \texttt{subbham@iitkgp.ac.in}
}

\begin{document}

\maketitle

\begin{abstract}
We propose an algorithm to explore the global optimization method, using SAT
solvers, for training a neural net. Deep Neural Networks have achieved great
feats in tasks like- image recognition, speech recognition, etc. Much of their
success can be attributed to the gradient-based optimisation methods, which scale
well to huge datasets while still giving solutions, better than any other existing
methods. However, though, there exist a chunk of learning problems like the parity
function and the Fast Fourier Transform, where a neural network using gradient-
based optimisation algorithm can’t capture the underlying structure of the learning
task properly (1). Thus, exploring global optimisation methods is of utmost interest
as the gradient-based methods get stuck in local optima. In
the experiments, we demonstrate the effectiveness of our algorithm against the
ADAM optimiser in certain tasks like parity learning. However, in the case of
image classification on the MNIST Dataset, the performance of our algorithm was
less than satisfactory. We further discuss the role of the size of the training dataset
and the hyper-parameter settings in keeping things scalable for a SAT solver.
\end{abstract}


\section{Introduction} 
\label{Chapter1} 


Machine Learning, at its core, is an optimisation problem. With highly non-linear models like neural networks, which have ushered a revolution in many fields of machine learning over the past decade or so, optimisation is still a challenging and non-trivial task. The state of the art optimisers for Neural Networks such as Adam, Adagrad and RMSProp  get stuck in spurious local optima. Finding a globally optimal solution is NP Hard. And to tackle this issue we leverage the prowess of SAT solvers which are highly engineered to find solutions to NP Hard Problems. The popular optimisation methods rely on a gradient based optimisation scheme and hence suffer from many drawbacks. For example, exploding and vanishing gradients tend to destabilise training and to mitigate this one has to resort to techniques like batch-normalisation. Including  a suitable regularisation strategy, they have a large number of hyperparameters to tune and finding a good combination of which is resource hungry and often requires intuition and experience.  Above all, these being greedy strategies we always end up with a sub-optimal solution. Thus we want need to move over the current trend of gradient based update methods and further strive to make neural network optimisation an elegant process with a minimal number of hyper-parameters to tune. In this paper we propose a non-greedy optimisation method to train a neural network featuring discrete weights. After bit-blasting the output of the neural network, we formulate the cost function as a Satisfiability problem a solution to which is found by using a SAT solver. As The weights and inputs being discretized we show how a modified version of relu called stepped-relu generalises better as an activation function. The number of hyper-parameters in our method is way lesser than the state of the art optimisers.  Our method being non-greedy finds a global minima wrt to the mini-batch of training examples. To make our algorithm scalable to huge datasets, we propose a method to parallelise training across batches of training data. Lastly a novel method to decrease the solving time of the sat solvers has been discussed.

\section{Related Works}
\label{relatedworks} 

The goal of this work is to design a non-gready optimisation scheme
for neural networks. To do this, the cost function of the neural network
is mapped to a SAT encoding and further SAT solvers were
used to find an assignment 
to the formula. Using such an encoding, Nina et al. \cite{narodytska2018verifying} explored 
various properties of a Binarized Neural Network
like robustness to adverserial perturbations. Furthermore,
Huang et al. \cite{huang2017safety} proposed 
a general framework for automated verification of safety of 
classification decisions made by feed-forward deep neural networks which 
leverages SMT solvers and SAT encodings. Zahra et al. \cite{ghodsi2017safetynets} proposed a 
framework that enables an untrusted server
(the cloud) to provide a client with a short mathematical proof of the correctness of
inference tasks that they perform on behalf of the client

\section{Reducing to Satisfiability} 
\label{Chapter3} 

\subsection{Satisfiability}

The Boolean satisfiability problem (sometimes called propositional 
satisfiability problem and abbreviated SAT) is the problem of determining 
if there exists an interpretation that satisfies a given Boolean formula. 
In other words, it asks whether the variables of a given Boolean formula 
can be consistently replaced by the values TRUE or FALSE in such a way 
that the formula evaluates to TRUE. For example a neural network model 
with let's say two variables only, $l_1$ and $l_2$, could look like -
$$\Sigma = (l_1 \vee l_2 \vee t_0) \wedge (\bar l_1 \vee l_2 \vee \bar t_0)$$
where $t_0$ is a temporary variable, created in the process of breaking the cost function
into CNF.
\subsection{Bit Blasting}

The primary idea here is to decompose the weight variables into bits and using binary arithmetics for addition, 
multiplication and non-linear activations reduce the cost function to a 0-1 optimisation problem. The weights 
bear a signed binary representation. A weight w is represented as,
\begin{equation}
  w = w_{0} w_{1} \dots w_{n-2} w_{n-1}, 
\end{equation}
where $w_{0}$ is the sign bit. 
The decimal value of w is then,
\begin{equation}
w_{decimal} = -2^{n-1}w_{0} + 2^{n-2}w_{1} \dots 2^{1}w_{n-2} + 2^{0}w_{n-1}
\end{equation}

\subsection{Signed and Unsigned Addition}

Addition of 2 signed numbers is done by \textit{carry look ahead adder}
method as shown in algorithm \ref{alg:bitwiseAdd} and setting the type to \textit{'signed'}.
As the sum has to be represented  in the same number of bits as addends’ , 
a constraint is generated which has to be satified for a valid addition.
For example, in the case of signed addition, the constraint 
is \textit{carryin} = \textit{carryout} wrt to the signed bit.
And for unsigned addition, the constraint is \textit{carryout}
from the signed bit = 0. These constraints are then converted to
CNF formula. We do so by using z3 solver.

\begin{algorithm}[tb]
\caption{Bitwise Addition: bitwiseAdd}\label{alg:bitwiseAdd}
\textbf{Input:} BitVectors: $a={a_{0} a_{1} \dots a_{n-2} a_{n-1}}$, $b={b_{0} b_{1} \dots b_{n-2} b_{n-1}}$, $type \in \{signed,unsigned\}$ \\
\textbf{Output:} BitVector $y = {y_{0} y_{1} \dots y_{n-2} y_{n-1}}$, with $y_{n-1}$ and 1 constraint
\begin{algorithmic}[1]

\STATE  {$n \gets length(a)$}
\STATE  {$carryPrev \gets 0$}
\STATE  {$carry \gets 0$}
\FOR   {$i \gets n-1$ to $0$} 
\STATE  {$carryPrev \gets carry$}
\STATE  {$G \gets a_i \wedge b_i$}
\STATE  {$P \gets a_i \vee b_i$}
\STATE  {$y_i \gets G\oplus P \oplus carryPrev$}
\STATE  {$carry \gets G \vee (P \wedge carryPrev)$}
\ENDFOR 

\IF{$type = signed$}
\STATE {$constraint \gets carry == carryPrev$}
\ELSE 
\STATE {$constraint \gets carry == 0$}
\STATE  \textbf{return} $y,constraint$
\ENDIF

\end{algorithmic}
\end{algorithm}

\subsection{Signed Multiplication}

First we convert the multiplicands into their magnitudes and 
then perform repeated unsigned addition by shift and add method.
Having found the magnitude, we assign a positive sign to it 
if the multiplicands are of the same sign, else a negative sign. 
The algorithm \ref{alg:bitwiseMul} illustrates the steps for this.

\begin{algorithm}[tb]
\caption{Bitwise Multiplication: bitwiseMul}\label{alg:bitwiseMul}
\textbf{Input:} BitVectors: $a={a_{0} a_{1} \dots a_{n-2} a_{n-1}}$, $b={b_{0} b_{1} \dots b_{n-2} b_{n-1}}$ \\
\textbf{Output:} BitVector $y = {y_{0} y_{1} \dots y_{n-2} y_{n-1}}$, with $y_{n-1}$ and constraints 
\begin{algorithmic}[1]

\STATE  {$n \gets length(a)$}
\STATE  {$a_{sign} \gets a_{0}$}
\STATE  {$b_{sign} \gets b_{0}$}

\STATE  {Initialise $product$ with \textit{slack\_bits} number of 0}
\STATE  {Initialise $a^{mag},b^{mag}$ with \textit{num\_bits} number of 0}

\FOR  {$i \gets 0$ to $n-1$}
\STATE  {$a_{i} \gets a_0\oplus a_i$}
\STATE  {$b_{i} \gets b_0\oplus b_i$}
\ENDFOR 

\STATE  {$a^{mag}_{n-1} \gets a_{sign}$}
\STATE  {$b^{mag}_{n-1} \gets b_{sign}$}

\STATE  a, constraint = bitwiseAdd(a, $a^{mag}$, signed)
\STATE  constraints $\gets$ \{constraint\}
\STATE  b, constraint = bitwiseAdd(b, $b^{mag}$, signed)
\STATE  constraints $\gets$ constraints $\cup$ \{constraint\}

\FOR   {$i \gets n-1$ to $0$} 
\STATE  {Initialise $b^{new}$ with \textit{slack\_bits} number of 0}
\FOR   {$j \gets 0$  to $n-1$}
\STATE  $b^{new}_{slack\_bits-1-j} \gets b_{n-1-j}\wedge a_i$
\ENDFOR 
\STATE  product, constraint = bitwiseAdd(product, $b^{new}$, unisgned)
\STATE  constraints $\gets$ constraints $\cup$ \{constraint\}
\ENDFOR 
\FOR   {$i \gets 0$ to $slack\_bits-product\_magnitude\_bits$}
\STATE  constraints $\gets$ constraints $\cup$ \{$product_i$ == 0\}
\ENDFOR 
\STATE  {$product_{sign} \gets a_{sign}\oplus b_{sign}$}
\STATE  {Initialise $product^{mag}$ with \textit{slack\_bits} number of 0}
\STATE  {$product^{mag}_{slack\_bits-1} \gets product_{sign}$}

\FOR  {$i \gets slack\_bits-1$ to $0$}
\STATE  {$product_{i} \gets product_{sign}\oplus product_i$}
\ENDFOR 

\STATE  product, constraint = bitwiseAdd(product, $product^{mag}$, signed)
\STATE  constraints $\gets$ constraints $\cup$ \{constraint\}

\STATE  \textbf{return} $y,constraints$

\end{algorithmic}
\end{algorithm}

The multiplicands are expressed in 2*\textit{num\_bits} number of bits.
The hyper parameter \textit{product\_magnitude\_bits} sets a limit
on the magnitude of the product. 
Ideally \textit{product\_magnitude\_bits} = 2*\textit{num\_bits}-1
But, we can set it to a lower value which would generate constraints
corresponding to product < $2^{\textit{product\_magnitude\_bits}+1}$.
This speeds up SAT solvers by limiting our search space to
a much smaller domain, with a compromise in the accuracy.

\subsection{Weighted Sum}

After the $w_i*x_i$ operation, we need to add all such products, which then
would be the input to one of the neurons in the next layer. 
Hence if the number of nodes in the previous layer is large, this raises a concern. 
Addition is done sequentially, this means there is an inherent upper bound on the partial sum. 
So, we introduce another hyper-parameter called \textit{slack\_bits}. Each term $w_i*x_i$ is sign extended 
from 2*\textit{num\_bits} to \textit{slack\_bits} number of bits. This mitigates the problem 
as now we can accommodate large temporary 
partial sums. Then this would mean that the input to the next layer 
would be in \textit{slack\_bits} number of bits. To prevent this, after doing the
weighted sum, we get rid of the least signigicant bits, to reduce the number of bits
from \textit{slack\_bits} to \textit{num\_bits}. In decimal, this corresponds to 
division with $2^{\textit{slack\_bits}-\textit{num\_bits}}$.
Note, unlike other weights, the bias is represented in \textit{slack\_bits} number of bits.

\subsection{Activation Function}

The hidden layer of the neural network features a non-linear activation function. We use 
Rectified Linear Unit (Relu) for this purpose. 
\[
relu(x)=  
\begin{cases} 
	x, & \text{if x $>$ 0.} \\
	0, & \text{otherwise.}
\end{cases}
\]
Lets see how relu activation function transforms the input bits into the output bits.
When the output $x <0$, $x_{n-1}$ the sign bit is 1. The output of relu should
be 0 (in decimal),i.e the output bits are set to 0. This is achieved by and-ing rest of the bits with MSB. Notice how the output is the same
as input when the input is a positive number i.e the sign bit is 0. So,

\begin{equation}
relu(x_{n-1} x_{n-2}\ \dots\ x_1x_0) = 0\ \overline{x_{n-1}}\wedge x_{n-2}\ \dots\ \\
\overline{x_{n-1}}\wedge x_1\ \overline{x_{n-1}}\wedge x_0
\end{equation}
Note the input to the activation function is in \textit{slack\_bits} number of bits. Because the output of this node 
further would be multiplied with a weight which is in \textit{num\_bits} number of bits, we want the output to be also 
represented in \textit{num\_bits} number of bits. 
Also we don’t want the output to blow up with each forward pass. Thus the activation function itself should take care 
of this. So if we want \textit{relu} as an activation function, we have to clip it’s maximum value at $2^{num\_bits-1}$-1. 
In this way the output can always be represented in \textit{num\_bits} number of bits with the MSB being the sign bit. 
The activation function is shown in figure and algorithm  \ref{alg:reluActivation} describes the operation.

\begin{algorithm}[tb] \caption{Activation Function: Relu}\label{alg:reluActivation}

\textbf{Input:} BitVector $x = {x_{0} x_{1} \dots x_{n-1} x_{n-1}}$\\
\textbf{Output:} BitVector $x = {y_{0} y_{1} \dots y_{n-1} y_{n-1}}$
\begin{algorithmic}[1]
\STATE  {$n \gets length(x)$}
\FOR  {$i \gets 1$ to $n-1$}
\STATE  {$y_i \gets x_i\wedge\overline{x_{n-1}}$}
\ENDFOR 
\STATE  {$y_{0}$ $\gets$ $0$}
\STATE  \textbf{return} $y$

\end{algorithmic}
\end{algorithm}

\begin{algorithm}[tb] \caption{Activation Function: Relu\_clipped}\label{alg:reluclipped}

\textbf{Input:} BitVector $x = {x_{0} x_{1} \dots x_{n-1} x_{n-1}}$
\textbf{Output:} BitVector $x = {y_{0} y_{1} \dots y_{n-1} y_{n-1}}$
and a list of constraints.
\begin{algorithmic}[1]

\STATE  {n $\gets$ length(x)}
\STATE  {temp $\gets$ Relu(x)}
\STATE  {temp $\gets$ $temp_{0} temp_{1} \dots temp_{n-1-regret\_bits}$}
\STATE  Prepend temp with 0s to make its length n
\FOR  {$i \gets 0$ to $n-1$}
\STATE  {$linearShiftDown_i \gets 1$}
\STATE  {$linearShiftUp_i \gets 0$}
\ENDFOR 

\FOR  {$i \gets n-num\_bits$ to $n-2$}
\STATE  {$linearShiftDown_i \gets 0$}
\ENDFOR 

\FOR  {$i \gets n-num\_bits$ to $n-1$}
\STATE  {$linearShiftUp_i \gets 1$}
\ENDFOR 

\STATE  {temp, constraintDown $\gets$ bitwiseAdd(temp, linearShiftDown, signed)}

\FOR  {$i \gets 1$ to $n-1$}
\STATE  {$temp_i \gets temp_i\wedge temp_{0}$}
\ENDFOR 

\STATE  {temp, constraintUp $\gets$ bitwiseAdd(temp, linearShiftUp, signed)}
\STATE  {constraints $\gets$ (constraintDown, constraintUp)}
\STATE  {y $\gets y_{slack\_bits-1-num\_bits} \dots y_{slack\_bits-1}$}
\STATE  \textbf{return} $y, constraints$
\end{algorithmic}
\end{algorithm}

In the above plot we see that due to clipping at 7, the output can be represented in 4 bits.
There is  one problem: Because the $w_i$ and $x_i$ are discrete, the input to a hidden node 
is very much susceptible to a slight change in $x_i$ which would hamper the generalising capability of 
the neural network. Hence to smoothen things out, we get rid of the last few least significant bits of the output, 
denoted as \textit{regret\_bits}, of the input and then apply the activation function. Getting rid of 
the last \textit{regret\_bits} bits has a effect of division with $2^{regret\_bits}$.  

\subsection{Architectures}
We propose two Neural Network Architectures namely-
\begin{itemize}
\item Vanilla Neural Network: The standard feedword network with relu activation function, a single hidden layer with 
10 hidden neurons and one single output node. 
\item Kernelised Neural Network: To take care of the problem as discussed in the discussion section, 
we want that the weights (in the first layer only) corresponding to the neighbouring pixels of the input image
should not vary significantly. Hence we can have a square matrix and a sliding window of size \textit{window\_size}. And 
the weights of the neural network are the average of the elements in the sliding window with stride \textit{window\_stride}.
\end{itemize}

\subsection{Cost Function}
We propose a cost function for a binary classification problem. The final output layer has a single neuron. The cost function, where y is the output: 
\begin{center}
$y \geq +2^{\textit{cost\_bits}}$ if $label = +ve$ \\
$y \leq -2^{\textit{cost\_bits}}$ if $label = -ve$ \\
\end{center}
Note that y is represented in \textit{slack\_bits} number of bits.
To implement the above cost function, we follow the algoithm \ref{alg:costFunction}.

\begin{algorithm}[tb]
\caption{Cost Function: cost}\label{alg:costFunction}
\textbf{Input:} BitVector $x = {x_{0} x_{1} \dots x_{n-1} x_{n-1}}$, $label \in \{-1,+1\}$\\
\textbf{Output:} constraints
\begin{algorithmic}[1]
\STATE  {$n \gets length(x)$}
\IF {y is +1}
\STATE  $constraints \gets \{x_0 == 0,\bigvee\limits_{i=1}^{n-cost\_bits-1}x_{i} == 1\}$
\ELSE 
\STATE  $constraints \gets \{x_0 == 1,\bigwedge\limits_{i=1}^{n-cost\_bits-1}x_{i} == 0\}$
\ENDIF
\STATE  \textbf{return} $constraints$
\end{algorithmic}
\end{algorithm}

\section{Implementation}
\label{implementation}

We start with a small experiment to check whether we can classify linearly separable points by training a perceptron model using this 
approach. The input was a 4 dimensional vector with each number being 0 or 1. A labeling function $y_{label} = sign(2x_0+3x_1-4x_2-2x_3+1)$
was chosen and used to label the points. Then the perceptron model described by $y = w_0x_0+w_1x_1+w_2x_2+w_3x_3+b$ was declared and the weights
$w_{0 \dots 3}$ and the bias are expressed with \textit{num\_bits = 4, product\_magnitude\_bits = 7, slack\_bits = 8}. Using the \textit{bitwiseAdd}
and \textit{bitwiseMul} method described as above, we express y in 8 bits. Setting the \textit{cost\_bits = 0} we get SAT. The dataset and 
assignments are given in ?.

\subsection{Dataset and Pre-processing}

\subsubsection{MNIST}
We run our experiments on MNIST Dataset. We want to learn a 3-recogniser. For this we first separate 3 and non-3 images.
Then create a train and test set of sizes 63 and 10139 with same number of both types. We follow 2 different pre-processing 
schemes for the vanilla and Kernelised Neural Network. The reasoning and details follow:
\begin{itemize}
\item Vanilla Neural Network: Downsampled the 28*28 image to 14*14 image.  
\item Kernelised Neural Network:  Borders with zero pixel values were 
	  removed while preserving the square structure of the image. Then downsampled to 10*10 image.
\end{itemize}

After pre-processing, the images were reshaped to a single dimensional form. Individual pixels (floating point number varying 
between 0 and 1) were discretized and represented in \textit{num\_bits} number of bits.  The following transformation gives 
the decimal representation of each pixel with value \textit{pixel\_val}:
\begin{center}
$K = int(pixel\_val*(2^{\alpha} -1))$ \\
\end{center}
where $\alpha = num\_bits - preprocess\_psub$ and $1 < \alpha < \textit{num\_bits}$.
Then K represented in \textit{num\_bits} number of bits, is the discretized value for that pixel. In our experiments we set $\alpha$
 = 2. It was observed that, with $\alpha >$ 2, the SAT solver couldn't find a solution to the cnf.

\subsubsection{Parity Learning}
Nye et al. \cite{nye2018efficient} show that gradient based optmisers
like \cite{kingma2014adam}, cannot be used to train a deep neural network
to learn the parity function. So, to test our algorithm we create 
two datasets of binary input vectors with $dimensionality \in \{8,16\}$.
To label every training example, we first randomly choose bit positions,
and then the label for that particular data becomes the xor of the
bits present in those positions. For example, for the dataset with 
dimensionality = 8, the positions chosen were 0, 1, 3, and 5. So,
the $i^{th}$ example $x_i$ has the label $y_i$ as,
$$y_i = x_i[0] \oplus x_i[1] \oplus x_i[3] \oplus x_i[5]$$
Similarly, for dimentionality = 16, the positions were 0, 1, 2, 3, 4, 6, 11, and 14.
\subsection{Generating CNF}

We use z3 solver to declare the weights of the neural network and its' libraries to do the binary arithmetic.
The forward pass of an image generates constraints (boolean equalities) because of addition, 
multiplication, activation functions and cost functions. The constraints for all the images in a 
given batch are fed to z3 solver, which then breaks down the complex expressions into cnf format. Let's denote
the generated clauses by $\Sigma_i$ for $i^{th}$ batch.

\subsection{Training}

From the train set of 63 images, we randomly sample 20 batches with \textit{batch\_size} 30. The training is divided into 2 phases-
\begin{itemize}
\item We generate $\Sigma_i$ for i: 1 $\rightarrow$ $num\_batch$. Then Using the following principle we collect implications from every batch. 
\begin{center}
$\Sigma_i \implies \overline{x_j} \vee x_k \vee \overline{x_l} \iff \Sigma_i \wedge {x_j \wedge \overline{x_k} \wedge x_l}$ is UNSAT
\end{center}
From above we conclude if $x_j, x_k$ and $x_l$ were assigned 1, 0 and 1 respectively in $\Sigma_i$ and $\Sigma_i$ becomes UNSAT,
then $ \overline{x_j} \vee x_k \vee \overline{x_l}$ is a learned clause.
Every implied clause carries information on how the weight variables are related among themselves to correctly classify that particular
batch. Let $\lambda_i$ be defined such as:
$$\lambda_i = \{l \mid \Sigma_i \Rightarrow l\}$$
$\lambda_i$ is generated in the following manner \ref{alg:ImpliedClauses}. Multiple batches are run parallely across different machines. 
One thing to note is time taken to find $\Sigma \wedge literals$ grows as \textit{varChunkSize} decreases. Because we don't want to get
stuck at one such initialisation, we choose to halt the process for a given assignment of \textit{literals} if we don't find UNSAT in
time = 180 seconds of run time. This number was chosen empirically.
\item After all the processes are complete in the previous step, we combine get 
$$\lambda^{all} = \bigwedge\limits_{i=1}^{num\_batches}\lambda_{i}$$ 
To find an assignment we run SAT solvers on all $\Sigma^{all}_i = \Sigma_i \wedge \lambda^{all}$ parallely across multiple machines.
\end{itemize}

\begin{algorithm}
\caption{Clause Sharing: ImpliedClauses}\label{alg:ImpliedClauses}
\textbf{Input:} Clauses $\Sigma$\\
\textbf{Output:} Implied Clauses $\lambda$
\begin{algorithmic}[1]
\STATE n $\gets$ length(vars)
\STATE varChunkSize $\gets$ n
\STATE Initialise $\lambda$=\{\}
\WHILE{getting learned clauses}
\STATE s $\gets \lceil \frac{n}{varChunkSize}\rceil$
\STATE count $\gets$ 0
\WHILE{count $<$ 100}
\FOR {$i \gets 1$ to s}
\STATE literals $\gets$ randomly sample varChunkSize no. of elements from vars
\STATE randomly set each element in literals to either 0 or 1
\IF {$\Sigma \wedge literals$ is UNSAT}
\STATE $\lambda \gets \lambda \cup \{\bigvee\limits_{j=1}^{varChunkSize}\overline{x_j}$ $\forall x_j \in literals\}$
\ENDIF
\ENDFOR
\STATE count $\gets$ count+1
\ENDWHILE
\STATE varChunkSize $\gets$ \textit{max}(varChunkSize - 0.05*varChunkSize, 50)
\ENDWHILE
\STATE \textbf{return} $\lambda$

\end{algorithmic}
\end{algorithm}

\subsection{Speeding Up computations in SAT Solver}

With the given choice of \textit{batch\_size} and architecture of the Network Network, the SAT solver could not find a solution 
to the $\Sigma^{all}_i$ in 48 hours of running the code. However things are boosted significantly by randomly setting a chunk of 
the model variables to either 0s or 1s and then running the solver. We start by assigning some 90\% model variables randomly to 
0s and 1s and follow the curriculum is shown in \ref{alg:assumptionSolving}. This also empowers us to find multiple solutions which 
are much different from each other. As \textit{varChunkSize} decreases, the likelihood
to find a solution and solving time both increase.

\begin{algorithm}
\caption{Solving: AssumptionSolving}\label{alg:assumptionSolving}
\textbf{Input:} Clauses $\Sigma$, model variables \textit{vars}\\
\textbf{Output:} Solutions to the Clauses $\lambda$
\begin{algorithmic}[1]
\STATE n $\gets$ length(vars)
\STATE Initialise sols=\{\}
\STATE solFound $\gets$ 0
\WHILE{solFound $<$ numSols}
\STATE count $\gets$ 0
\WHILE{count $<$ 100}
\FOR {$i \gets 0$ to s}
\STATE literals $\gets$ randomly sample varChunkSize no. of elements from vars
\STATE randomly set each element in literals to either 0 or 1
\IF {$\Sigma \wedge literals$ is SAT}
\STATE assignment $\gets$ solution to $\Sigma \wedge literals$
\STATE $sols \gets sols \cup$ \{assignment\}
\STATE solFound $\gets$ solFound+1 
\IF {$size(sols) = numSols$}
\STATE \textbf{return} $sols$
\ENDIF
\ENDIF
\ENDFOR
\STATE count $\gets$ count+1
\ENDWHILE
\STATE varChunkSize $\gets$  \textit{max}(varChunkSize - 0.05*varChunkSize, 50)
\ENDWHILE
\STATE \textbf{return} $sols$
\end{algorithmic}
\end{algorithm}

\section{Results}
\label{results}

Experiments were done to see how various parameters and model architectures 
influence the qualitiy of a solution found by the Sat Solver. For experiments in \ref{regretBits},\ref{slackBits},
\ref{costBits}, \ref{ModelArchitecture} each instance of the solver was run on a 8 core cpu with hyper-threading,
16 GB RAM and  4 plingeling solver threads. The experiments in \ref{KernelParams} and \ref{SharingandNoSharing}
were run on a 24 core machine with hyperthreading, 32GB RAM and 12 plingeling threads as finding a solution was
much harder in these cases. For all the experiments num\_bits = 4, \textit{product\_magnitude\_bits} = 7 and a 
neural network with a single hidden layer with 10 nodes were used unless otherwise stated.

In table \ref{regretBits}, we set \textit{slack\_bits} = 8, \textit{cost\_bits} = 3, and observe that increasing
\textit{regret\_bits} improves generalising capacity of the neural network. \textit{slack\_bits} bottle-neck 
the weighted sums. For example, if the input node is of 196 dimensions, 
hidden nodes in the subsequent layer receive the weighted sum of 196 numbers. With \textit{slackBits} set to 8, we impose a constraint
such which enforces partial sums to be small enough to be stored as a 8 bit fixed-point number. To study the effect of \textit{slackBits},
in table \ref{slackBits}, we set \textit{cost\_bits} = 4, we find that increasing \textit{slack\_bits}
doesn't improve accuracy. Because $regret\_bits \leq slack\_bits-num\_bits$, with more \textit{slack\_bits} we could
vary the \textit{regret\_bits} too. And we see that increasing $regret\_bits$ doesn't make accuracy any better.

\begin{table}[t]
\caption{Variation with \textit{regret\_bits}}
\label{regretBits}
\begin{center}
\begin{tabular}{c|c|c|c}
\textit{regret\_bits} 	&Min 		&Median	 &Max 
\\ \hline
 & & &\\
2		&47.98\%	&60.93\%		&74.62\%\\
3		&51.79\%	&66.36\% 		&77.15\%\\
4		&41.48\%	&67.18\%		&82.04\%\\
\end{tabular}
\end{center}
\end{table}

\begin{table}[t]
\caption{Variation with \textit{slack\_bits}}
\label{slackBits}
\begin{center}
\begin{tabular}{c|c|c|c|c}
\textit{slack\_bits}	&\textit{regret\_bits}	&Min 		&Median	 &Max 
\\ \hline
 & & & &\\
8		&4	&53.46\%  	&71.08\% 	&82.81\%\\
9		&5	&58.62\%  	&74.37\% 	&80.11\%\\
10		&6	&61.00\%  	&71.87\% 	&80.51\%\\
\end{tabular}
\end{center}
\end{table}

In table \ref{costBits}, with \textit{regret\_bits} = 4, \textit{slack\_bits} = 8, we realise 
even with stronger separations between $y_{pred}^{+}$ and $y_{pred}^{-}$ the accuracy doesn't go up.
In table \ref{ModelArchitecture}, we study if increasing the complexity of the neural network improves the accuracy. We set 
\textit{regret\_bits} = 4, \textit{cost\_bits} = 4, \textit{slack\_bits} = 8. The accuracy still doesn't improve.

\begin{table}[H]
\caption{cost\_bits}
\label{costBits}
\begin{center}
\begin{tabular}{c|c|c|c}
Type			&Min	&Median	&Max
\\ \hline
 & & &\\
3	&41.48\%  &67.18\%  &82.04\%\\
4	&53.46\%  &71.08\%  &82.81\%\\
\end{tabular}
\end{center}
\end{table}

\begin{table}[H]
\caption{Variation with model architecture}
\label{ModelArchitecture}
\begin{center}
\begin{tabular}{c|c|c|c|c}
Layers		&Hidden nodes		&Min			&Median 	&Max 
\\ \hline
 & & & &\\
1	  		&10		&53.46\%	&71.08\% 	&82.81\%\\
1	 		&20		&56.56\%	&68.06\%	&78.14\%\\
2	  		&5, 5	&61.75\%	&72.52\%	&78.75\%\\
\end{tabular}
\end{center}
\end{table}

In table \ref{SharingandNoSharing}, \textit{regret\_bits} = 4, \textit{cost\_bits} = 4, \textit{slack\_bits} = 8, 
by looking at the min and median scores we conclude that the clause sharing across batches is not effective at all. Also 
seeing twice as more data doesn't improve the max accuracy.

\begin{table}[t]
\caption{Clause sharing vs no sharing vs all}
\label{SharingandNoSharing}
\begin{center}
\begin{tabular}{c|c|c|c}
Type			&Min	&Median	&Max
\\ \hline
 & & &\\
Clause Sharing	&52.46\%  &69.73\%  &81.75\%\\
No Sharing		&53.46\%  &71.08\%  &82.81\%\\
Entire dataset	&70.19\%  &75.24\%  &80.64\%\\
\end{tabular}
\end{center}
\end{table}

\begin{figure}
	\centering
	\includegraphics[width=4cm,height=3cm,angle=0]{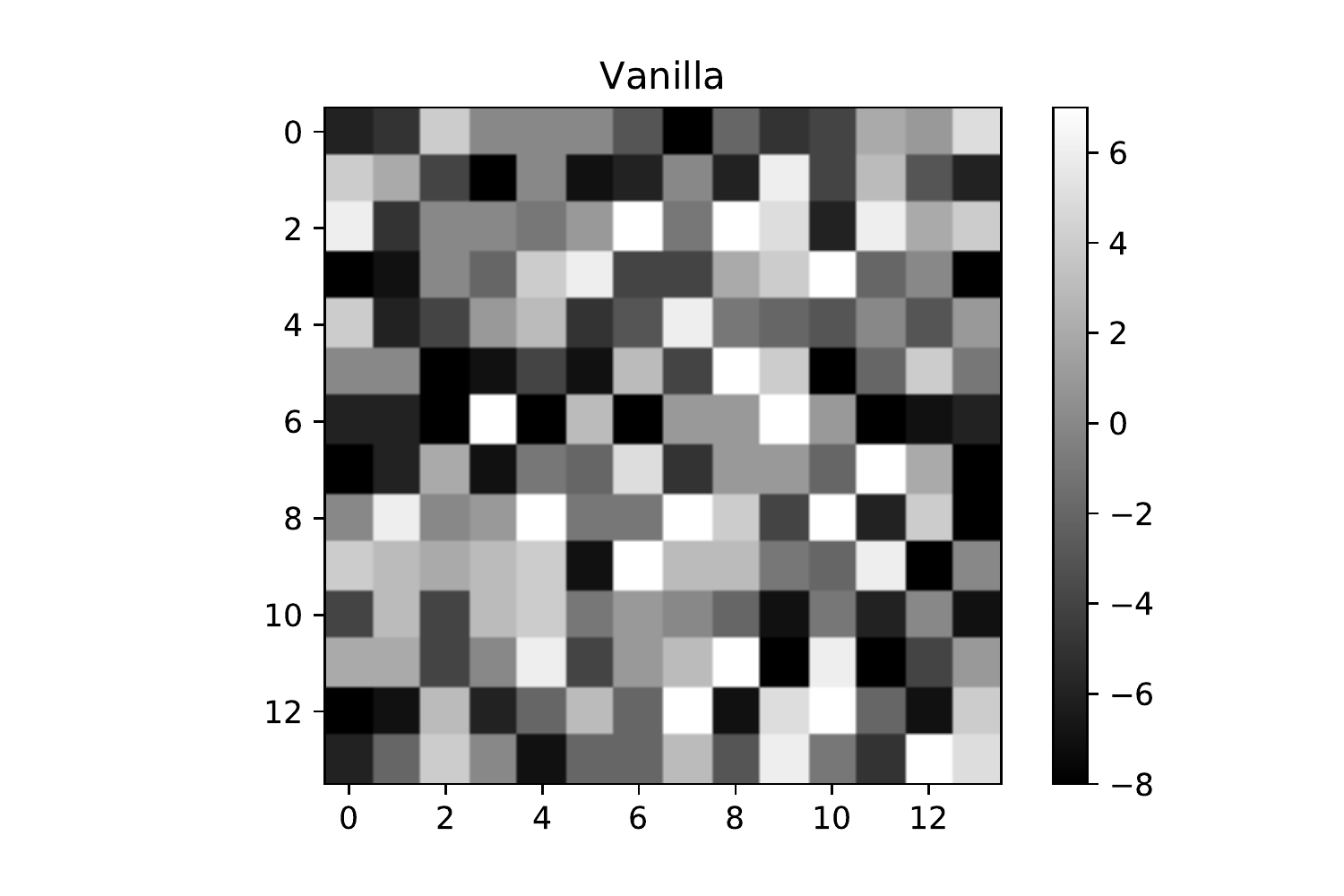}\label{img:vnn}
	\includegraphics[width=4cm,height=3cm,angle=0]{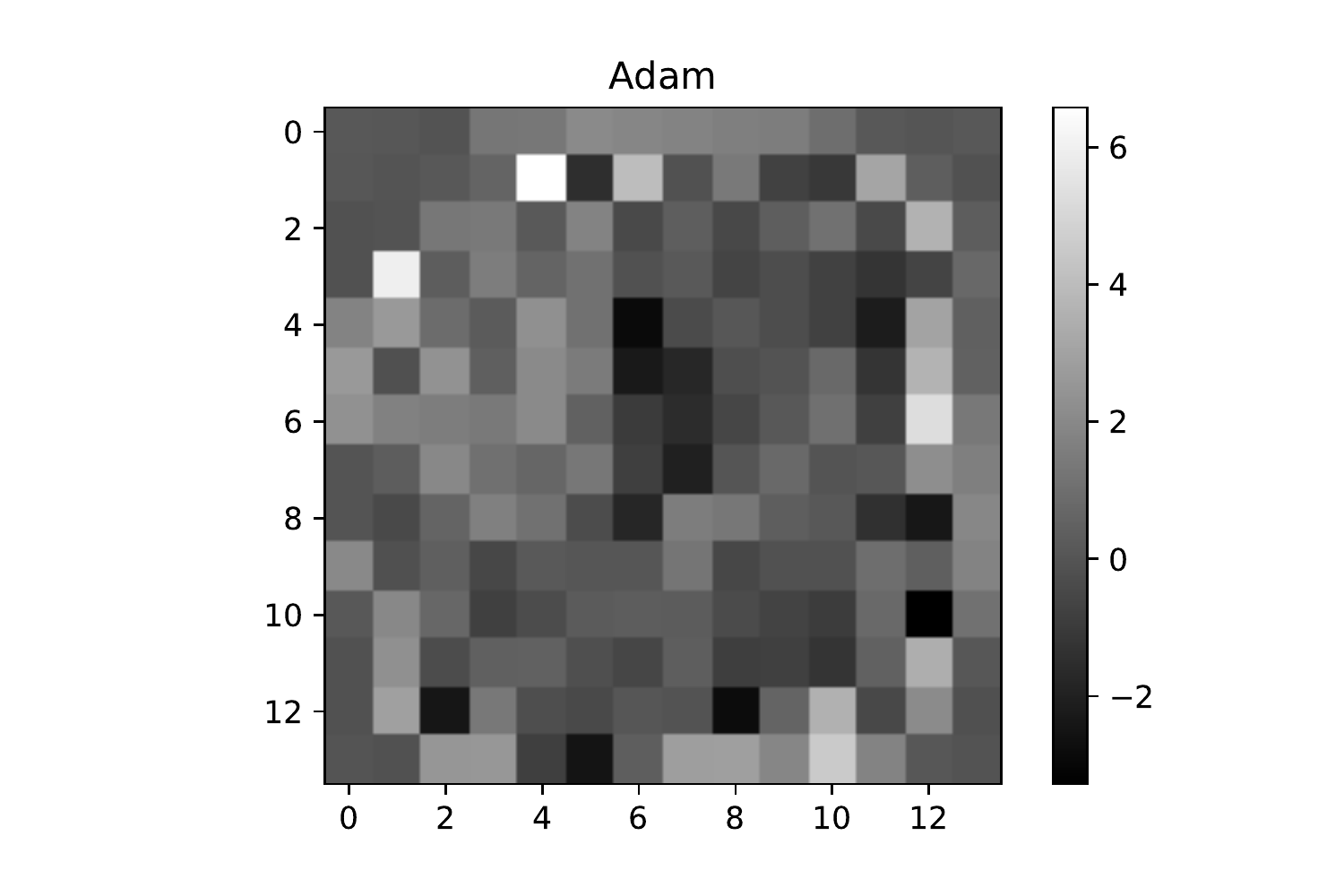}\label{img:adam}
	\includegraphics[width=3.3cm,height=2.14cm,angle=0]{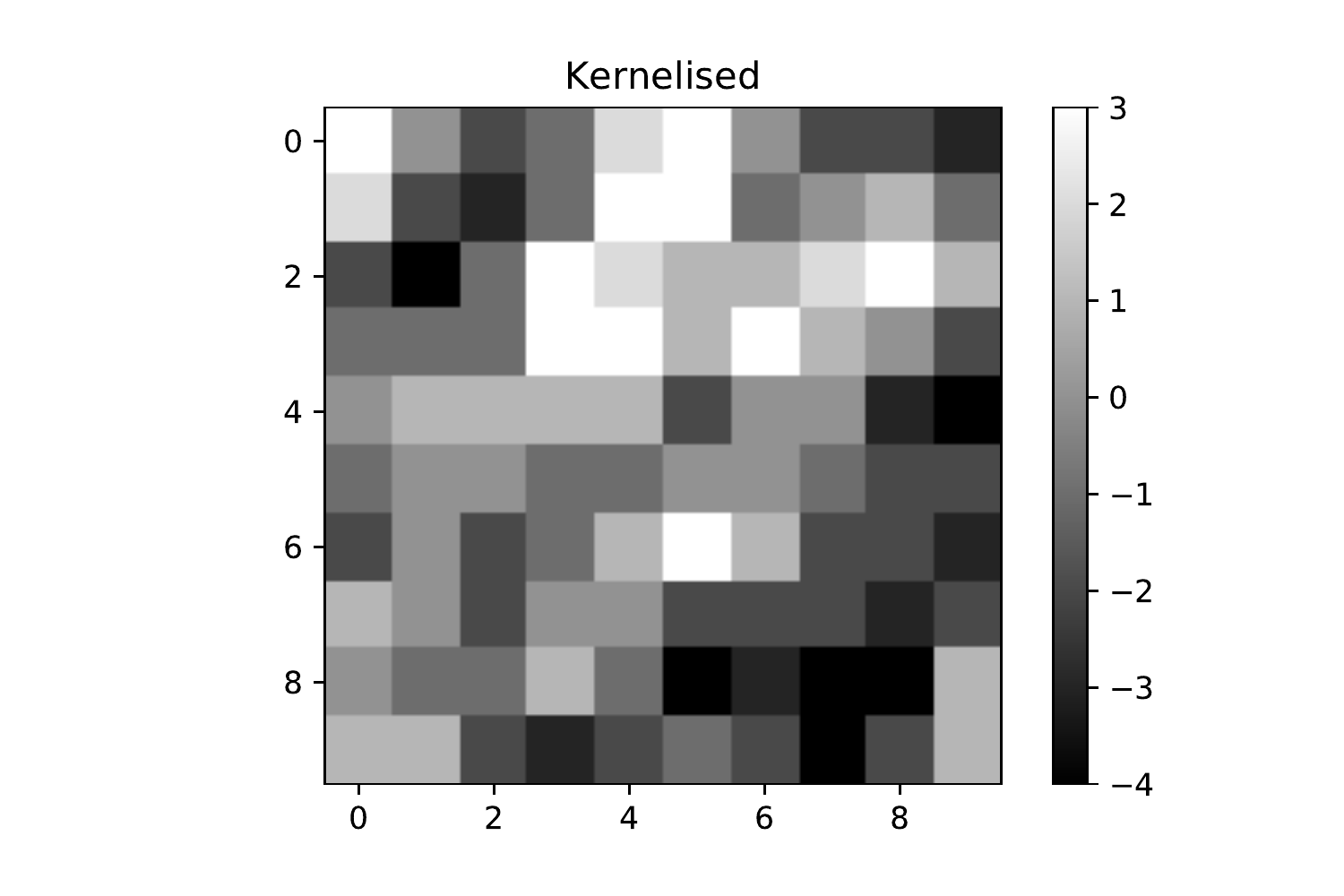}\label{img:knn}
	\caption{Plot of the weights in the 1st layer, from the input to a hidden node.}
	\label{fig:gridWeights}
\end{figure}

To get a better understanding of things hampering the performance, look at the fig \ref{fig:gridWeights}

\begin{figure}
	\centering
	\includegraphics[width=4cm,height=4cm,angle=0]{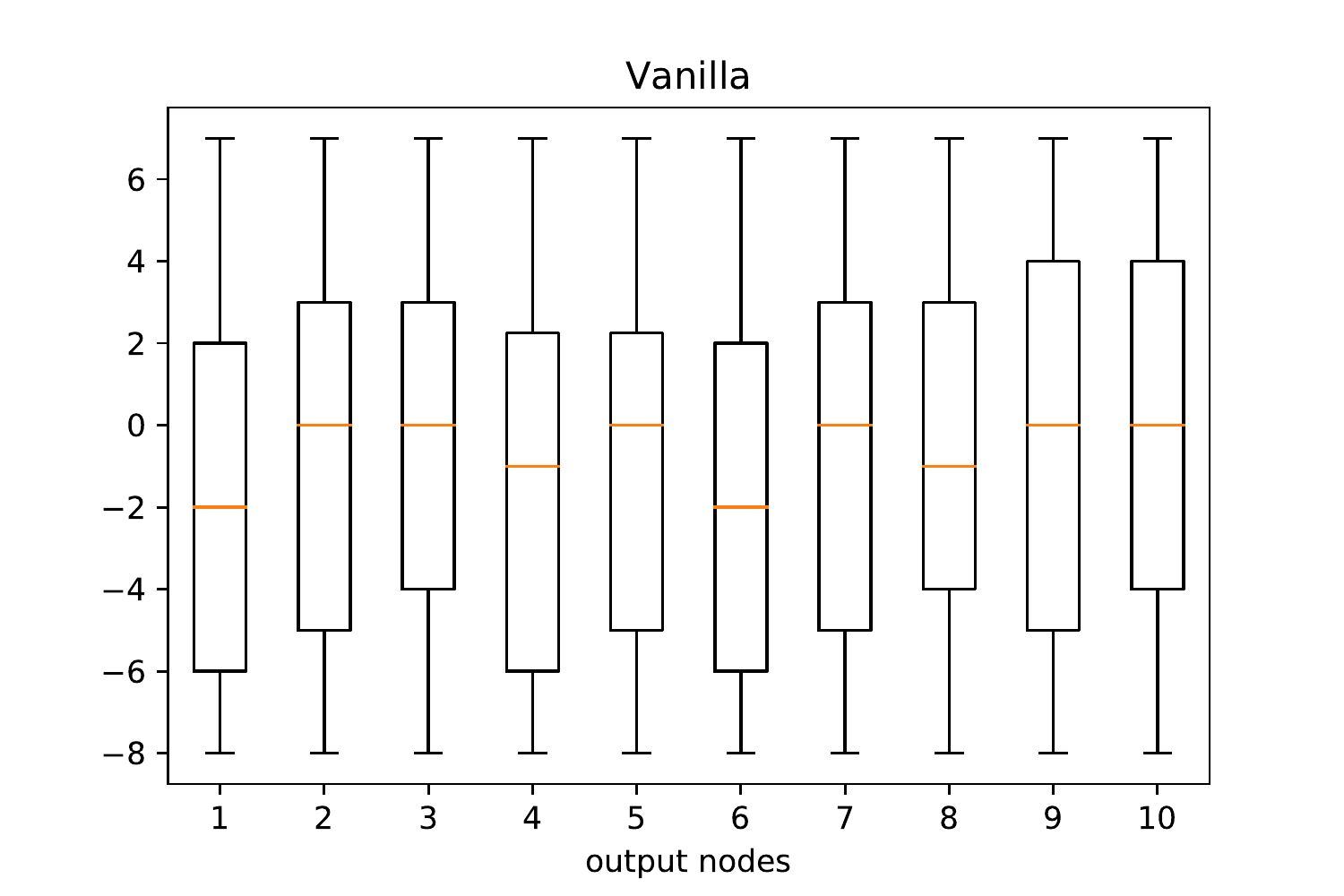}
	\includegraphics[width=4cm,height=4cm,angle=0]{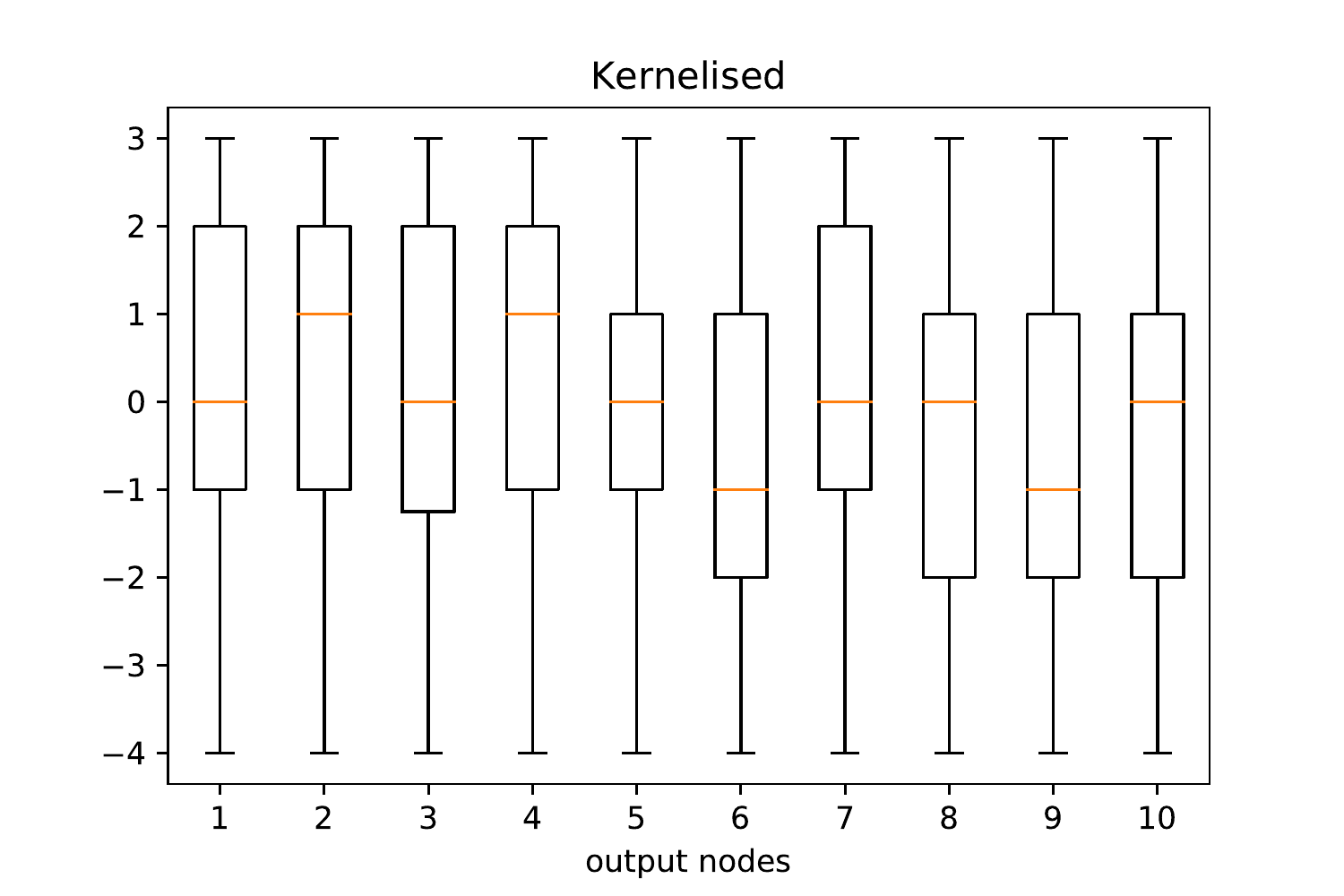}
	\includegraphics[width=4cm,height=4cm,angle=0]{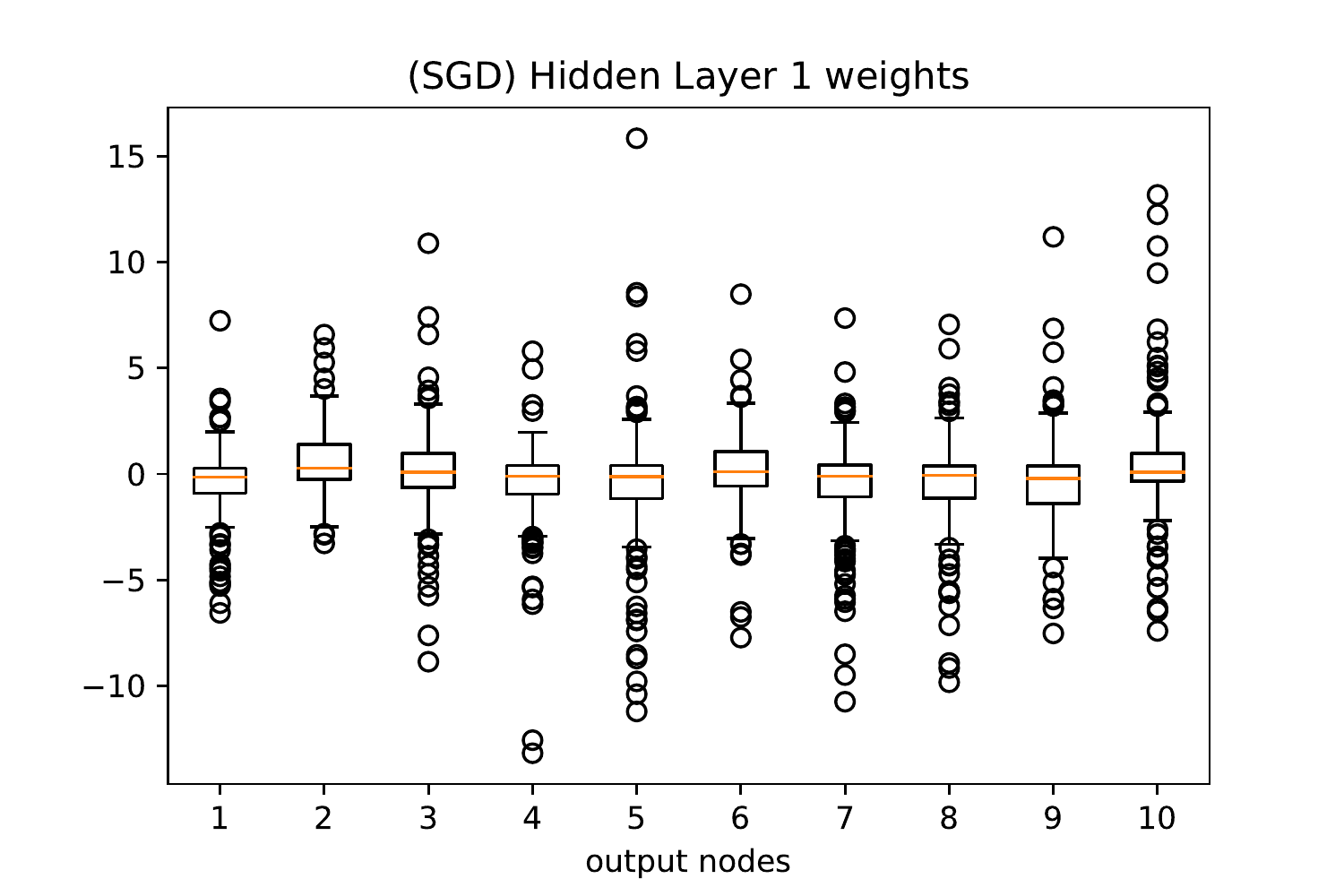}
	\caption{Distribution of the weights in the 1st layer.}
	\label{fig:weightsBoxPlot}
\end{figure}

As we observe, the problem is not the architecture of the network but the kind of weights that are being learnt.
Above we see that the weights in the first layer learnt by the sat solver are quit arbit as 
compared to the ones learnt by adam optimiser. Probably this is the reason we are not able to 
increase the accuracy. This is the main motivation for \methodKNN. Now that we have weights 
that are moving window averages, the neighbouring weights do not vary much. We set  
\textit{slack\_bits} = 8, \textit{kernel\_stride} = 2, \textit{kernel\_rb} = 1 and run the experiments
in \ref{KernelParams}

\begin{table}[H]
\caption{Variation with Kernel params}
\label{KernelParams}
\begin{center}
\begin{tabular}{c|c|c|c|c}
\textit{kernel\_size}	&\textit{cost\_bits}	&Min	&Median	&Max
\\ \hline
 & & & &\\
3		&3		&73.96\%  &76.89\% 	&80.07\%\\
3		&4		&77.37\%  &77.57\% 	&78.02\%\\
4		&3		&61.91\%  &78.02\%	&81.78\%\\ 
4		&4		& - 	& - 	& - \\ 
\end{tabular}
\end{center}
\end{table}

As stated earlier, we make random assumptions to the weight bits and feed the cnf to the sat solver. 
This might be the reason we learn bits that are really arbit. But then again it is equally likely for a 
Sat solver to find any assignment as long as the cnf is SAT. So not making any prior assignment would not 
guarantee a solution where the learnt weights are not co-related among neighbouring pixels.

In case of Parity Learning, we see our algorithm outperforms gradient descent by a significant margin. 
With gradient descent the training accuracy remains around ~50\%. CITE This being a binary classification
task, we could infer that gradient descent doesn't make any useful updates to the neural network. 
But, our algorithm not only achieves a \textbf{100\%} train accuracy but also \textbf{100\%} train accuracy
for the 8 bit and \textbf{57.50\%} accuracy in case of the 16 bit dataset.  

\begin{table}[H]
\caption{Binarized Neural Network}
\begin{center}
\begin{tabular}{c|c|c|c|c}
\textit{batch\_size} &\textit{cost\_bits}	&Min	&Median	&Max
\\ \hline
& & & &\\
20	&1	&46.81\%  &59.20\%	&69.38\%\\
20	&2	&45.55\%  &59.29\%	&69.34\%\\
20	&3	&55.76\%  &64.91\%	&76.21\%\\ 
20	&4	&67.16\%  &71.59\%	&77.56\%\\ 
30	&1	&64.80\%  &68.65\%	&77.31\%\\
30	&2	&66.13\%  &76.07\%	&\textbf{83.10}\%\\
\end{tabular}
\end{center}
\end{table}

\begin{table}[H]
\caption{XOR (clause sharing)}
\begin{center}
\begin{tabular}{c|c|c|c}
Bits 	&Min	&Median	&Max
\\ \hline
& & &\\
8	&62.50\%  &87.50\%	&\textbf{100.00}\%\\
16	&40.00\%  &50.00\%	&\textbf{57.50}\%\\
\end{tabular}
\end{center}
\end{table}

\section{Discussion}
\label{discussion}
In this work, we have presented a non-greedy
optimisation scheme to train neural networks. 
Despite being non-greedy, 
we find in tasks like image classification
gradient descent based optimisers outperform our 
algorithm. This can be attributed to the fact that
our algorithm doesn't scale to the point where
we see the entire dataset. 
Due to the very small amount of data we see, our algorithm 
overfits easily. For our future work, we would like to
explore more methods to parallelise learning across batches
so that the net training error improves.
\bibliography{main}

\begin{thebibliography}{5}
\providecommand{\natexlab}[1]{#1}
\providecommand{\url}[1]{\texttt{#1}}
\expandafter\ifx\csname urlstyle\endcsname\relax
  \providecommand{\doi}[1]{doi: #1}\else
  \providecommand{\doi}{doi: \begingroup \urlstyle{rm}\Url}\fi

\bibitem[Narodytska et~al.(2018)Narodytska, Kasiviswanathan, Ryzhyk, Sagiv, and
  Walsh]{narodytska2018verifying}
Nina Narodytska, Shiva Kasiviswanathan, Leonid Ryzhyk, Mooly Sagiv, and Toby
  Walsh.
\newblock Verifying properties of binarized deep neural networks.
\newblock In \emph{Proceedings of the AAAI Conference on Artificial
  Intelligence}, volume~32, 2018.

\bibitem[Huang et~al.(2017)Huang, Kwiatkowska, Wang, and Wu]{huang2017safety}
Xiaowei Huang, Marta Kwiatkowska, Sen Wang, and Min Wu.
\newblock Safety verification of deep neural networks.
\newblock In \emph{International conference on computer aided verification},
  pages 3--29. Springer, 2017.

\bibitem[Ghodsi et~al.(2017)Ghodsi, Gu, and Garg]{ghodsi2017safetynets}
Zahra Ghodsi, Tianyu Gu, and Siddharth Garg.
\newblock Safetynets: Verifiable execution of deep neural networks on an
  untrusted cloud.
\newblock \emph{Advances in Neural Information Processing Systems}, 30, 2017.

\bibitem[Nye and Saxe(2018)]{nye2018efficient}
Maxwell Nye and Andrew Saxe.
\newblock Are efficient deep representations learnable?
\newblock \emph{arXiv preprint arXiv:1807.06399}, 2018.

\bibitem[Kingma and Ba(2014)]{kingma2014adam}
Diederik~P Kingma and Jimmy Ba.
\newblock Adam: A method for stochastic optimization.
\newblock \emph{arXiv preprint arXiv:1412.6980}, 2014.

\end{thebibliography}








\end{document}